\documentclass[sigconf]{acmart}

\usepackage{booktabs} 
\usepackage{enumitem}
\usepackage{url}
\usepackage{setspace}

\newif\ifdraft\drafttrue

\ifdraft

\newcommand\todo[1]{{\footnotesize \color{red}[#1 - \textbf{TODO}]}}
\newcommand\tf[1]{{\footnotesize \color{green}[#1 - \textbf{Tobi}]}}
\newcommand\mz[1]{{\footnotesize \color{blue}[#1 - \textbf{Martina}]}}
\newcommand\hjc[1]{{\footnotesize \color{yellow}[#1 - \textbf{Hyung Jin}]}}
\newcommand\ac[1]{{\footnotesize \color{blue}[#1 - \textbf{Antoine}]}}
\newcommand\mpet[1]{{\footnotesize \color{magenta}[#1 - \textbf{Max}]}}
\newcommand\yd[1]{{\footnotesize \color{red}[#1 - \textbf{Yiannis}]}}
\else
\newcommand\tf[1]{}
\newcommand\mz[1]{}
\newcommand\hjc[1]{}
\newcommand\ac[1]{}
\newcommand\mpet[1]{}
\newcommand\yd[1]{}
\newcommand\todo[1]{}
\fi

 \copyrightyear{2018} 
 \acmYear{2018} 
 \setcopyright{acmlicensed}
\acmConference[GECCO '18]{Genetic and Evolutionary Computation Conference}{July 15--19, 2018}{Kyoto, Japan}
 \acmPrice{15.00}
 \acmDOI{10.1145/3205455.3205571}
 \acmISBN{978-1-4503-5618-3/18/07}

\begin{document}
\title{Hierarchical Behavioral Repertoires with Unsupervised Descriptors}

\author{Antoine Cully}
\orcid{0000-0002-3190-7073}
\affiliation{%
  \institution{Imperial College London}
}
\email{a.cully@imperial.ac.uk}

\author{Yiannis Demiris}
\affiliation{%
  \institution{Imperial College London}
}
\email{y.demiris@imperial.ac.uk}

\begin{abstract}
Enabling artificial agents to automatically learn complex, versatile and high-performing behaviors is a long-lasting challenge. This paper presents a step in this direction with hierarchical behavioral repertoires that stack several behavioral repertoires to generate sophisticated behaviors. Each repertoire of this architecture uses the lower repertoires to create complex behaviors as sequences of simpler ones, while only the lowest repertoire directly controls the agent's movements. This paper also introduces a novel approach to automatically define behavioral descriptors thanks to an unsupervised neural network that organizes the produced high-level behaviors. The experiments show that the proposed architecture enables a robot to learn how to draw digits in an unsupervised manner after having learned to draw lines and arcs. Compared to traditional behavioral repertoires, the proposed architecture reduces the dimensionality of the optimization problems by orders of magnitude and provides behaviors with a twice better fitness. More importantly, it enables the transfer of knowledge between robots: a hierarchical repertoire evolved for a robotic arm to draw digits can be transferred to a humanoid robot by simply changing the lowest layer of the hierarchy. This enables the humanoid to draw digits although it has never been trained for this task.

\end{abstract}
%
%



\maketitle

\section{Introduction}

\begin{figure}[!t]
\centering \includegraphics[width=1\columnwidth]{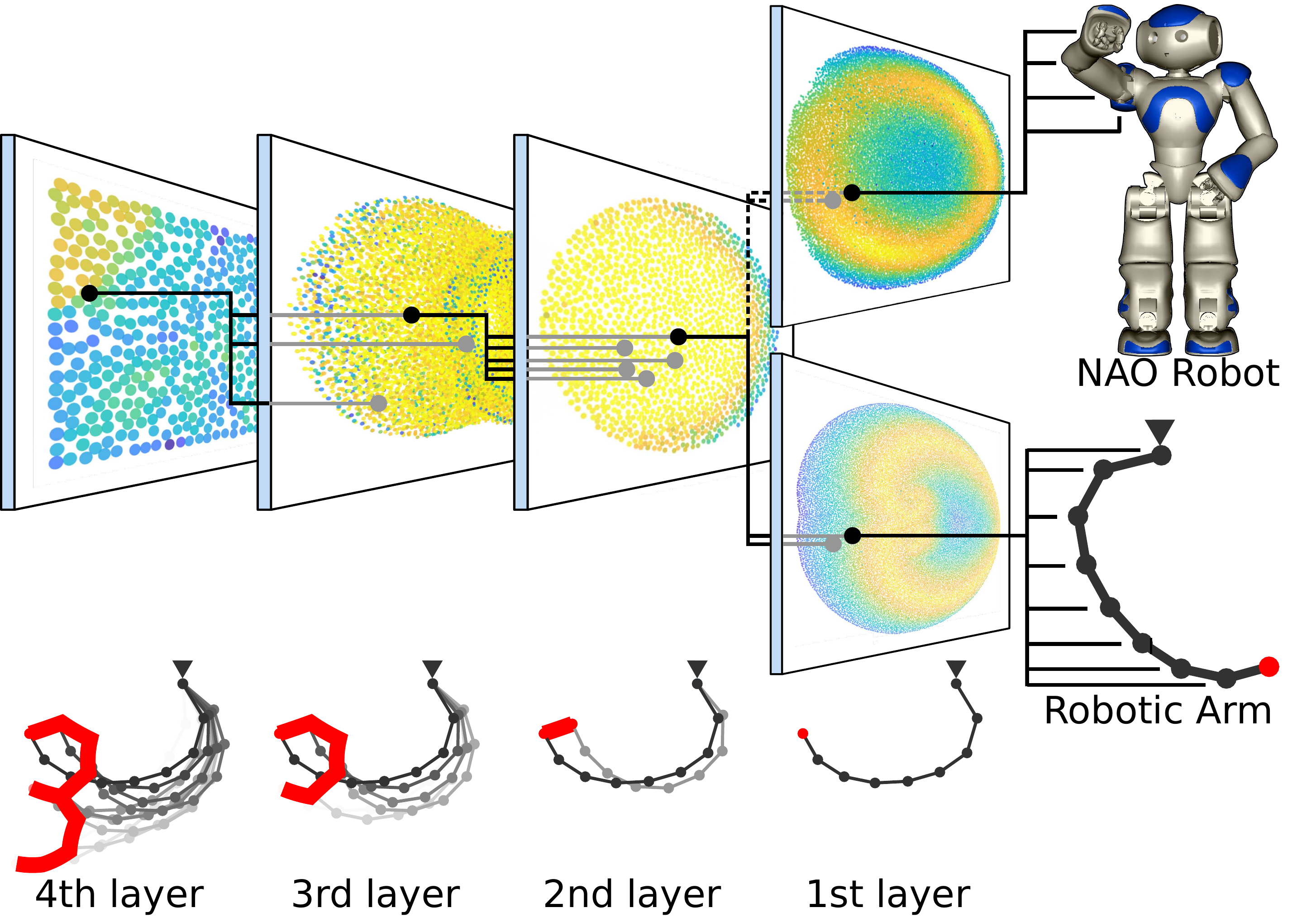}
\caption{A Hierarchical Behavioral Repertoire (HBR) is a hierarchical organization of behavioral repertoires that select controllers from lower levels of the hierarchy to form complex behaviors. Only the lowest layer of the hierarchy controls the robot's movements. This architecture increases the scalability of BRs, while reducing the dimensionality of the optimization task. The learned behaviors can be transferred to other robots by changing the bottom layer of the hierarchy, as denoted with the dashed connections.}
\label{fig:concept}
\end{figure}

The complexity, versatility, and fitness of organisms produced by natural evolution fascinate many computer scientists whose ambition is to transfer these features to artificial systems~\citep{eiben2015evolutionary, doncieux2015evolutionary}. In particular, the evolutionary robotics field already offered several exciting results in this direction~\citep{doncieux2015evolutionary}, for examples, with the automatic generation of morphologies and controllers of real robots~\citep{lipson2000automatic}, or by providing the ability to overcome mechanical damages to physical robots~\citep{bongard2006resilient,cully2015robots}.

Recently, researchers have proposed the concepts of Behavioral Repertoire (BR)~\cite{cully2013behavioral,duarte2017evolution} evolution and Quality-Diversity (QD)~\cite{cully2017quality,pugh2016quality} optimization that generate sets of diverse and high-performing behaviors (or optimized solutions). These approaches have shown to outperform traditional evolutionary approaches thanks to better exploration capabilities~\citep{ cully2013behavioral, pugh2015confronting, cully2015robots}. In robotics, using a large set of diverse behaviors extends the versatility of robots and enables them to face many situations. Unfortunately, we can not increase the versatility of robots with BR by simply adding more and more dimensions to the repertoire. This likely leads to an exponential growth of the repertoire size, which might become intractable, dilute the selective pressure, and thus deteriorate the quality of the produced solutions.

Several studies have observed that the modularity, and the hierarchy of the brain neuronal structure plays an important role in the complexity and versatility of natural organisms~\cite{zhou2006hierarchical, wagner2007road}. 
The objective of this paper is to bring the benefits of modularity and hierarchy to the domain of BR and QD optimization. The proposed Hierarchical Behavioral Repertoires (HBRs, see Fig.~\ref{fig:concept}) leverage the diversity of controllers from the repertoires in the different levels of the hierarchy to build complex behaviors, while maintaining the overall complexity of the architecture tractable.

We illustrate the capabilities of the HBRs by enabling a robotic arm and a humanoid robot to learn how to draw digits in an unsupervised fashion after having learned how to draw lines and arcs. The ability to draw sequences of arcs and lines is used to approximate arbitrary trajectories, as explained in \citep{cully2015evolving}.
Conversely to supervised learning or imitation tasks, in which the robot tries to reproduce motor trajectories, our robots are never presented with examples of motor trajectories that generate a digit. The robots instead learn by trying different motor strategies until finding a trajectory that generates a shape which looks similar to images from a dataset of handwritten digits. This scenario is similar to a child learning how to write by looking at examples of digits in their textbook. An illustrative video is available here: \url{https://youtu.be/maSr8DR1uh8}.

The experimental results show that the proposed HBRs offer instrumental features to generate versatile, complex and high-performing BRs:\\\begin{minipage}[t]{1\linewidth}\vspace{-2mm}
\begin{enumerate}[label=\textbf{\arabic*}),wide, labelwidth=!, labelindent=0pt]
\item Deep neural networks can be used to automatically define the behavioral descriptor in QD-optimization algorithm in an unsupervised way. 
\item Only the lowest layer of the architecture is specific to the robot. Thus, the abilities learned by one robot can be instantaneously transferred to another robot by connecting the upper layers of the architecture to a first layer designed for the new robot.
\item The proposed architecture breaks down challenging problems into tractable sub-problems. In practice, it reduces the size of the repertoire by several orders of magnitude. 
\item In our experiments, the average quality of the behaviors was 2.3 times better than traditional approaches.
\item It provides inherent constraints to the generated behaviors, which make them less prone to fool the neural networks used to evaluate the behaviors.

\end{enumerate}
\end{minipage}

\section{Related Work}
\subsection{Behavioral Repertoire Evolution}
A BR is a set of simple controllers that governs the motor commands sent to the robot. The controllers are organized according to a behavioral descriptor, which characterizes the behavior of the robot while executing this controller. 
For instance, the behavioral descriptor can represent the x/y location of a robot after executing a controller for a few seconds. Such a descriptor has been used to enable a hexapod robot to learn how to walk in every direction~\citep{cully2013behavioral, cully2015evolving}. 

The original BR-evolution paper~\citep{cully2013behavioral} suggested that BRs can be combined with a high-level algorithm, like a planning algorithm, selecting controllers to steer the robot in the desired directions.
This approach has been followed by~\cite{chatzilygeroudis2018reset} to enable a robot to face damage situations, while avoiding obstacles.
A similar idea has been used in \cite{duarte2017evolution}, in which the authors evolved a neural network that selects behaviors from a BR to control a hexapod robot in a maze. The neural network selects the behavioral descriptor, and thus the associated controller, that should be executed according to the data from the robot's sensors (e.g., wall detectors). 

One limitation of BRs is that a different controller is required for each of the behaviors that the robot may have to execute. The consequence of this is that the size of the BR rapidly increases up to a point of being intractable. For instance, walking in every direction means to have a controller for all the possible angles, and speeds. If we want also to integrate another aspect in the descriptor, like a stability margin, the number of controllers will grow exponentially, as dictated by the curse of dimensionality. The HBRs presented in this paper are specifically designed to enable a larger diversity of behaviors in the BR while maintaining the size of the BR tractable.

\subsection{Quality-Diversity Optimization}
Instead of searching for a single high-performing solution like most optimization algorithms, QD optimization~\citep{pugh2016quality, cully2017quality} aims to produce a large collection of solutions that are both diverse and high-performing. QD algorithms are particularly instrumental to generate BRs as they enable the generation of large collections of different controllers covering the range of possible behaviors for robots, while each of them being high-performing.
 
The Novelty Search with Local Competition (NSLC)~\citep{lehman2011evolving, cully2013behavioral} and MAP-Elites~\citep{mouret2015illuminating} are the two main QD-algorithms. They have shown to be particularly instrumental in the field of evolutionary robotics, for instance, by allowing robots to overcome mechanical damages~\citep{cully2015robots}, or to evolve complex neural networks for maze navigation~\citep{pugh2015confronting}. Several variants of these two main algorithms have been proposed using different containers~\citep{vassiliades2017using, smith2016rapid}, and selection operators~\citep{mengistu2016evolvability, gravina2016surprise, gaier2017data}.
A unifying framework has been proposed to gather these different variants into a common formalism~\citep{cully2017quality}. This new representation enabled the definition of different operators and containers, and has shown that several combinations of operators and containers outperform MAP-Elites and NSLC in multiple tasks. 

While any QD-algorithm can be used to evolve HBRs, we will use the archive-curiosity variant from the framework described above~\cite{cully2017quality}, which combines an unstructured archive (similar to the one in NSLC) with a selection operator based on a ``curiosity score''. This score fosters the selection of individuals that are likely to produce offspring that will be added to the archive, and has been shown to outperform competing approaches on several benchmarks, including a robotic arm similar to the one used in this paper~\cite{cully2017quality}. 

\subsection{Innovation Engine}
In this paper, we will combine QD algorithms with deep neural networks for the automatic determination of the behavioral descriptors. 
This idea was initially introduced through the concept of Innovation Engines~\citep{nguyen2015innovation, nguyen2016understanding}. One of the strengths of deep neural networks is their ability to automatically extract features for the recognition of objects from complex data, like images or videos~\citep{lecun2015deep}. Innovation engines use this ability to automatically associate the behavioral descriptor of the solutions created by the QD-algorithms to the labels predicted by the network. The confidence of the network in its predictions is used as the fitness function to foster the QD-algorithm to generate solutions that can be recognized without ambiguity.  

For instance, this approach has been used to generate artificial images using hyper-NEAT~\cite{stanley2009hypercube} with an Alex-net trained in a supervised way to recognize objects from the ImageNet dataset. The QD-algorithm has been able to generate images looking like strawberries, TVs, donuts and many other examples~\citep{nguyen2015innovation}. However, the same team also pointed-out that neural networks can be easily fooled by the artificial images generated by the QD-algorithm~\citep{nguyen2015deep}. The algorithm generated synthetic images totally unrecognizable to human eyes, while being recognized by the network with a confidence larger than $99.6\%$. We show that the hierarchical structure of the proposed architecture can generate solutions that are less likely to fool deep neural networks.

\section{Methods}
The objective is to enable BRs to generate complex behaviors via the combination of two new mechanisms: 1) Hierarchical Behavioral Repertoires (HBR), which form a hierarchy of BRs, and 2) the automatic determination of behavioral descriptor by using neural networks trained in an unsupervised fashion.

\subsection{Hierarchical Behavioral Repertoire}
The first mechanism involves using several BRs that are stacked in a hierarchical architecture (see Fig~\ref{fig:concept}). With this approach, instead of using a planning algorithm or a neural network for the ``high-level'' algorithm, as suggested in the previous BR papers~\citep{cully2013behavioral, duarte2017evolution, chatzilygeroudis2018reset}, here we propose to use another BR to select and combine behaviors from the first BR. 
With this configuration, the high-level repertoire does not send motor commands to the robot, but rather successively calls behaviors from the low-level repertoire. These combinations of low-level behaviors form high-level and more complex behaviors.  
The architecture is evolved by successively generating each layer of the hierarchy starting from the first layer and climbing up through the hierarchy.

This concept of hierarchical BRs can be employed with more than two layers. In this paper, we will use up to four layers to enable our robots to learn how to draw points, lines, arcs, and digits respectively (see Fig~\ref{fig:concept}). However, in our experiments we did not notice any practical limitation that prevents more layers to be used, while four layers were sufficient to fulfill the tasks.

The incremental generation of the different layers of the hierarchy shares some links with the domain of incremental evolution~\cite{mouret2008incremental}, which uses intermediate tasks~\cite{gomez1997incremental}, or parametrized fitness functions~\cite{nolfi1995evolving} to enable the evolutionary process to bootstrap on complex scenarios. In particular, behavioral decomposition~\cite{larsen2005evolving} also aims to learn a hierarchy of primitive behaviors and arbitrators to solve complex tasks.
However, it is important to recall that the BRs contain thousands of different behaviors that are generated and organized automatically, to form a continuous  behavioral space, that can be useful to learn other behaviors (e.g., dance moves) or adapt to unforeseen situations (e.g., damage situations~\cite{cully2015robots}).

One challenge with HBRs is that chaining behavior executions make them start from arbitrary initial conditions. The produced behaviors might thus perform differently, as they might depend on the state the robot reached after executing the previous behavior. For instance, a controller designed to draw a 2cm long line, will be unable to properly execute this behavior if the robot starts next to the border of its reachable space. 

A naive approach to solve this issue is to extend the behavioral descriptor with information about the robot's and environment's state. 
However, this approach increases the length of the behavioral descriptor and the BR might quickly become too large to be tractable (the size of the BR being exponentially proportional to the dimension of the descriptor space). In the experiments of this paper, we show that generating a BR with this configuration becomes a real challenge in terms of computational requirements with poor performance in general.

In order to overcome this difficulty, we propose to use a stochastic behavioral descriptor to characterize the probabilistic distribution of the robot's behavior. 
The stochastic behavioral descriptor is defined as the geometric median (which is an extension of the traditional median for multidimensional data \citep{lin1992approximation}) of the traditional descriptor after uniformly sampling the initial configuration of the robot. In practice, during evaluation, the controller is evaluated several times (e.g., 100 times) with different initial conditions. For each of these evaluations, the traditional behavioral descriptor is recorded. When all the evaluations have been performed, the geometric median of the descriptors is computed. This stochastic behavioral descriptor is then used as a traditional descriptor to place the individual in the repertoire.

\begin{figure}
\centering \includegraphics[width=\columnwidth]{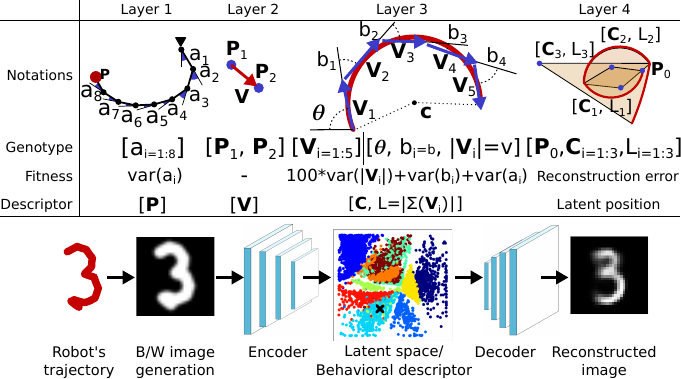}
\caption{(Top) Behavioral descriptor, genotypes, phenotypes, and fitness used for each layer (see section \ref{sec:exp_details} for details). (Bottom) Processing a digit trajectory through the autoencoder in order to extract its latent position and reconstruction error. }
\label{fig:notation}
\end{figure}

The average distance between the geometric median and the behavioral descriptor of each evaluation can be computed and used as a quality metric that quantifies the uncertainty of the stochastic descriptor. If the average distance is close to $0$, this means that the behavior is almost deterministic. In the other case, if the average distance is large, this means that the behavioral descriptor is largely dependent on the initial conditions. This metric can be used by the controllers of higher-levels of the hierarchy. For instance, a controller could try to minimize the uncertainty associated with the selected behavioral descriptor to minimize the variability of the produced behavior. This is not investigated in this paper and is left for future work. 
The fitness associated with the solution is implemented as the fitness of the median individual.

It is important to note that not all the layers require a stochastic behavioral descriptor. Certain layers can be independent from the initial conditions of the system. For instance in our experiment, only the second and third layers use stochastic behavioral descriptors.

\begin{figure*}[!t]
\centering \includegraphics[width=0.95\textwidth]{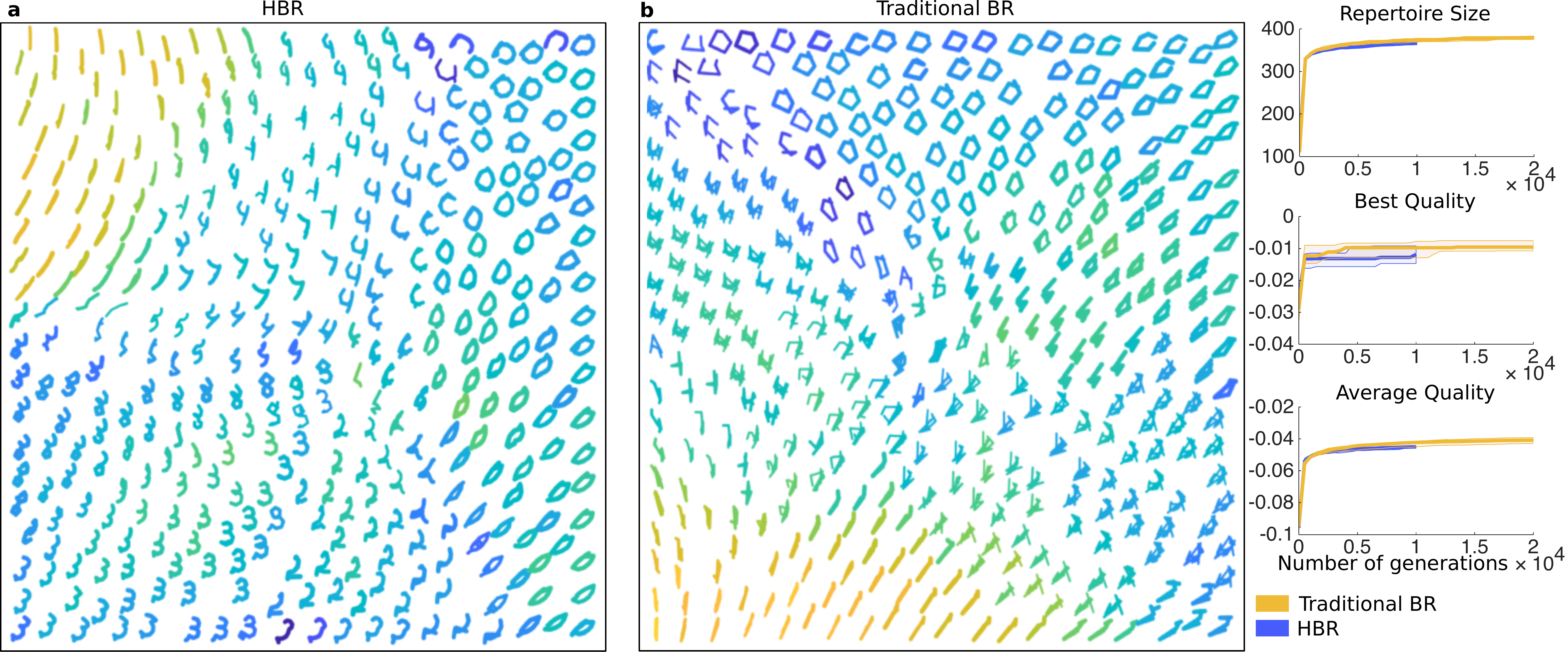}
\caption{a) Digits drawn by the robotic arm using the HBR. b) Digits drawn using a single BR layer, with directly encoded controllers. Right: BR size, best quality and average quality over the BR for each of the 10 replications. The solid lines represent the median while the shaded areas extend to the first and third quartile. }
\label{fig:main_res}
\end{figure*}

\subsection{Automatic Behavioral Descriptors}
While behavioral descriptors are manually defined in the vast majority of the literature \citep{cully2017quality, pugh2015confronting, duarte2017evolution, mouret2015illuminating}, we demonstrate that a deep neural network trained in an unsupervised fashion can be used to automatically define them.
In particular, we use this approach to define the descriptors of the last layer of the hierarchy, as it is particularly challenging to manually determine a single expression that characterizes all the 10 digits. Thanks to this neural network, our robots become able to autonomously reproduce elements from an image dataset of handwritten digits. In our future work, we will investigate the possibility to extend this concept to all the layers of the hierarchy.

The combination of a QD algorithm and a deep neural network has already been explored in \cite{nguyen2016understanding,nguyen2015innovation,nguyen2015deep}. However, in all those cases, the neural network can only be trained in a supervised manner. Here, we propose to use a convolutional deep autoencoder, which enables the construction of a behavioral descriptor in a fully unsupervised manner. This is an important feature as it can be challenging to define clusters or labels in certain datasets. For instance, this can be used to reproduce the motions of another (observed) robot, without the need for human intervention.

An autoencoder is composed of two sub-networks, an encoder network and a decoder network. The role of the autoencoder is to jointly train the sub-networks so that they learn how to encode the input data with the first sub-network and to reconstruct it via the second one. Autoencoders have many applications and variants, like denoising autoencoder, variational autoencoder or contractive autoencoder~\citep{Goodfellow-et-al-2016}. In this paper, the feature that mainly interests us is the ability of autoencoders to reduce the dimensionality of the input data. We use it to automatically project high-dimensional data from the observed behaviors (e.g., trajectories, sensor data) into a low dimensional representation that can be used as a behavioral descriptor. In our experiments, the autoencoder is used to transform the images of handwritten digits (784 pixels) from the MNIST dataset into a 2D latent space (constrained between $[-1, 1]$). The details of the autoencoder architecture, as well as the parameter values used in the QD algorithm, are given in appendix~\ref{ap:nn}.

Fig.~\ref{fig:notation}-bottom illustrates the entire pipeline from the robot trajectory to the behavioral descriptor: the trajectory of the gripper is recorded and then transformed into a black and white image (similar to the MNIST dataset). The image is then encoded and the 2D location of the image in the latent space is recoded. This location is used as the behavioral descriptor for the trajectory. Finally, the image is decoded and the ``reconstruction error'' (i.e., the pixel-wise mean squared error between the original and the decoded image) is computed and used as the fitness value of the trajectory. This fitness function promotes behaviors that generate trajectories and images that are similar to those in the MNIST dataset.

\section{Experimental Scenario}
In this paper, the main objective of the HBR is to allow a robot to execute arbitrary trajectories. Thus, the objective of the three first layers of the HBR is to enable our robots to execute a large diversity of lines and arcs from different initial configurations. Later, we will show how these three layers can be used to build a fourth layer that enables our robots to learn how to reproduce handwritten digits in an unsupervised manner.

\subsection{Robots}
The first robot used in this paper is a 8 degrees of freedom (simulated) planar robotic arm similar to the one used in \cite{cully2015robots, cully2017quality}. Its 8 joints are controlled in angular position mode, and the $X$/$Y$ position of its gripper is monitored. 
In the second experiment of this paper, we use a simulated NAO humanoid robot with 25 degrees of freedom. In our experiment, we will only consider the 4 joints of the right arm, which are controlled in angular position mode. The $X$/$Y$/$Z$ positions of the hand are recorded.

\subsection{Global Architecture}
The evolved HBRs are composed of four layers. The first layer enables the robots to reach every point in their reachable space. The generation of this layer follows closely the algorithms and methods used in \cite{cully2015robots, cully2017quality}. The second layer enables the robots to draw lines of different lengths and directions by connecting two points from the first layer. The behaviors of the first layer that are in between these two points are executed to create smooth transitions (see the supplementary video).
The third layer combines five lines from the second layer to empower the robot with the ability to draw arcs of different lengths, radius and orientations. Finally, the last layer grants the robots the ability to draw digits by concatenating arcs and lines from the lower layers. 
The definitions of the behavioral descriptor, genotypes, phenotypes, and fitness used for each layer are described in Fig.~\ref{fig:notation}-top. Additional implementation details are given in the following subsections.

\subsection{Layer 1 to 4: points, lines, arcs, and digits}\label{sec:exp_details}
The three first layers use common principles for the behavioral descriptors, genotypes, phenotypes, controllers and fitness functions. 
The behavioral descriptor of each layer characterizes the sub-trajectory produced by the controller. It either captures the position $\textbf{P}$ (layer 1), the linear displacement $\textbf{V}$ (layer 2) or the drawn arc defined by the position of its center $\textbf{C}$ and its length $L$ (layer 3, see Fig.~\ref{fig:notation}-top).  
The three layers use a directly encoded phenotype, which represents a set of parameters for an open-loop controller. The genotypes are defined as a set of float values defined between $0$ and $1$. The phenotypes map the genotype values into the ranges of the controller parameters. In particular, the range of the joint angular positions in Layer 1 are between $-\pi/2$ and $\pi/2$ radians for the robotic arm. For the higher layers, the ranges are defined to match the behavioral descriptor space of the layers below. 

The fitness function of the first layer is defined as the variance of the joint positions (which needs to be minimized). This fitness encourages every joint to contribute equally to the motion, which is important for smooth transitions between two configurations. No fitness function is defined for the second layer (the objective of the QD algorithm is only to cover the range of possible behaviors). 
The fitness function of the third layer captures the regularity of the generated arcs. This is achieved by minimizing the variance of the segments' length ($\mid\textbf{V}_i\mid$) and the variance of the angles between each segment ($b_i$). An additional term is used in the fitness function to minimize the variance of the joint positions (like the fitness of the first layer). This term is added for a fair comparison with reference algorithms as explained below. The sum of the three terms is roughly weighted with a factor $100$ on the length variance to make the influence of each term uniform (see Fig.~\ref{fig:notation}-top). 

We also propose an alternative implementation of the third layer's controller, in which only three parameters are needed (instead of ten, used to define the five lines).
We use this second implementation for the generation of the architecture with four layers for simplicity. However, the first implementation of this controller is used for the quantitative evaluations involving only three layers (explained later in the text). 
The second implementation guarantees that the lines will follow a circular trajectory and minimize the fitness function defined above. 
In this second implementation, the first parameter governs the direction (between $[-\pi; \pi]$) of the first segment. The second parameter defines the length of all the segments to make them equal. The last parameter controls the angle ($[-\pi; \pi]$) between each segment.

The controller used in the last layer is similar to the ones of the first layers, except that the length of the genotype can grow or contract to enable the robot to control the number of arcs and lines used in the trajectory. The two first parameters of the genotypes/phenotypes define the initial position of the robot (based on Layer 1). The third parameter defines the line width of the drawing. The rest of the genotype is composed of a set of triplets. The number of triplets is between one and three and is governed by the mutation operator ($5\%$ of chance to increase or reduce the number of triplets). Each triplet encodes an arc from Layer 3. The set of triplets forms the sequence of arcs and lines that define the complete trajectory of the robot.

\section{Experimental Validation}

\subsection{From motor babbling to drawing digits}\label{sec:exp_numbers}
In this first experiment, we use the proposed HBR to enable our robotic arm to learn how to draw digits, while starting from motor babbling. This experiment illustrates the ability of the HBR to evolve behaviors of increasing complexity through the layers of the hierarchy. 
In particular, we compare the quality of the digits generated by the HBR with similar digits generated by a traditional BR\cite{cully2013behavioral}.
This single layer architecture uses between one to three sets of five configurations (which corresponds to a total of between 40 and 120 parameters) to encode the trajectory. This genotype is designed to have the same capabilities as the genotypes used with HBR. The experiment has been replicated 10 times (including the training of the autoencoder). 

In Fig.~\ref{fig:main_res}-a, we can see the digits generated by a HBR. The numbers are organized according to their location in the latent space (i.e., the behavioral descriptor of layer 4). We can see that most of the digits are represented and that there is a continuum between the different classes. For instance, the 3s progressively turn into 8s, which then turn into 4s, 7s and 1s. In future works, this property could enable robots to discover innovative solutions that are not contained in the training dataset of the autoencoder. 
We can also note that certain classes of digits are less represented than the others (for instance the 6s in Fig.~\ref{fig:main_res}-a). This is due to the difficulty for the auto-encoder to distinguish certain types of digits.
In the future, we will investigate the use of other deep neural architectures or dimensionality reduction techniques (like kernel principal component analysis) to mitigate these inaccuracies.

Fig.~\ref{fig:main_res}-b shows the numbers generated by the traditional BR. While the latent space is similarly covered, we can see that the quality of the produced digit is particularly degraded, even with twice more generations. Apart from the 0s and 1s, which are reasonably well executed, all the other digits look like scribbles barely resembling a digit. More surprisingly, when we compare the highest and average quality of the BR according to the autoencoder (i.e., the reconstruction error), the traditional BR outperforms the HBR, while we can observe the opposite (the two presented replications are randomly selected and reflect the general tendency of the results).
We argue that this is another example that neural networks can easily be fooled~\citep{nguyen2015deep}. Indeed, in this experiment, there are no ``adversarial examples'', like in generative adversarial networks~\citep{wang2017magan}, to train the network to reject the scribbles produced by the robot. Conversely, the network is looking for features that are specific to each digit (e.g., specific angles, orientations, or intersections). The reference algorithm has more freedom than the HBR to explore the search space and to produce the exact feature the network is looking for, even if the final result is not convincing from a human perspective.

It is interesting to see that the HBR is less affected by the weaknesses of the network than the traditional BR. We hypothesize that this is the result of inherent constraints imposed by the different layers that condition the type of trajectory our robot is able to produce. For instance, encouraging the trajectory to contain arcs and not only a succession of abrupt angles, is very likely one of the reasons explaining the higher quality of the digits produced by our architecture.  

The quantitative evaluations in the next sections only rely on the three first layers (hard-coded fitness functions), in order to avoid the confusion provided by the autoencoder.

\begin{figure}
\centering \includegraphics[width=\columnwidth]{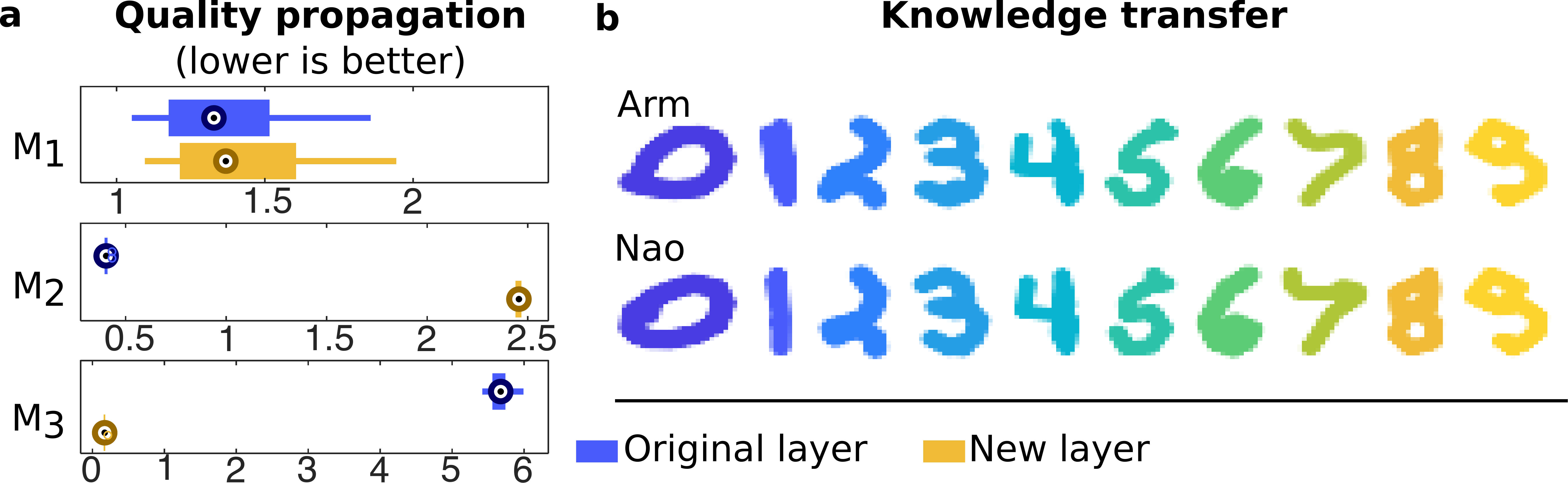}
\caption{a) Influence of substituting the first layer of the architecture with a layer using a different fitness function. The b) Digits generated by the robotic arm and the humanoid NAO robot.}
\label{fig:transfer}
\end{figure}

\subsection{Quality propagation and knowledge transfer}
In this second experiment, we show two important properties of the proposed architecture: 1) the ability to propagate the quality of solutions from a layer through the upper-layers of the architecture, and 2) the possibility to transfer the knowledge of our robotic arm to another robot. These two properties are the direct consequences of the hierarchical and modular organization of the HBR and can have significant applications in the domains of robotics and artificial intelligence. 

First, we show we can substitute layers of the hierarchy with BRs using different fitness function as long as the behavioral descriptor definition remains the same. The new fitness function propagates through the upper layers as their behaviors become instantaneously aligned with this new function.
In the experiment, we generate a new Layer 1, which instead of minimizing the variance between the angular positions of the robot's joints, encourages the robot to use only two of its joints for the motion. This new fitness function is defined as the sum of the 6 smallest squared angular positions of the robot's joints (among 8 joints). Minimizing this function fosters the robot to use two joints to execute most of the movement.

To evaluate the effect of substituting the first layer of the hierarchy without further evolutionary steps, we re-evaluate the solutions contained in Layer 3. For each solution, we record three metrics: ($M_1$) the two first terms of the fitness function of Layer 3 (those that encode the quality of the arcs), ($M_2$) the last term of the fitness function of Layer 3, which corresponds to the fitness of the original Layer 1, and ($M_3$) the fitness of the new Layer 1.

The boxplots in Fig~\ref{fig:transfer}-a show that using the original Layer 1 or the new one does not affect the first metric: the quality of the arcs is preserved regardless of the layer used. Conversely, we can see that substituting the first layer changes the two other metrics: when using the new Layer 1, the solutions of the third layer provide the best results regarding $M_3$, while becoming less effective with respect to $M_2$. 
This result illustrates that changing one layer enables the robot to instantaneously switch the proficiency of the hierarchy, without further evolution of the system. For instance, this can be used to generate multiple running modes, like a safe-mode and a dynamical-mode. Switching between these two modes can easily be done by switching from one layer to another. 

This concept can be extended to transfer knowledge between different robots. Instead of generating a layer 1 with a different fitness, we generate a Layer 1 designed for a different robot (see Fig~\ref{fig:concept}). As long as the new layer entirely overlaps the region of the behavioral descriptor covered by the original layer, the rest of the hierarchy can be directly used with the new robot without any adaptation. We applied this principle with a NAO robot. We generate a Layer 1 that corresponds to a 2D slice of its 3D reachable space (the behavioral descriptor has been scaled to cover the same area as the original layer 1). 

To evaluate the similarity between the digits drawn by NAO of those from the arm, we generate the image of the trajectory for each solution in Layer 4 (with both robots). Then, we compute the average pixel-wise difference between the image generated by NAO and those generated by the arm. The results show that  the difference is lower than $1\%$ over the pixels of the image (median 0.92\%[ min 0.10\%; max 2.11\%] excluding outliers and considering all the images of the 10 runs together). This is further demonstrated in  Fig~\ref{fig:transfer}-b, where we can see that it is challenging to see differences between the two robots. 
It may happen that some behaviors do not transfer properly. An example is visible in the supplementation video, as the NAO robot is unable to reproduce a 4. This comes from slight differences between the coverage of the two first layers. We expect that fine tunning (e.g., via further training) the entire architecture with the new robot can quickly remove these rare issues. 

\subsection{The benefits of stochastic descriptors}
For this experiment, we use the two first layers of the hierarchy (Layers 3 and 4 are not used) to quantify the influence of the stochastic descriptor on Layer 2. As a reference, we consider the same two layers, except that the second layer uses a traditional descriptor that is extended to include the initial state of the robot. This extension of the behavioral descriptor is feasible in this case because the resulting descriptor contains only four dimensions (two for the initial position of the gripper, and two for its displacement). Considering larger behavioral descriptors would be particularly challenging, as explained above. The first layer of the architectures follows the same definition and evolution procedure for both approaches, only the second layer is different.

In order to investigate whether the number of samples used for the computation of the stochastic descriptors has an influences on the results, we consider four variants of HBR using different numbers of samples and total number of generations: 1) 1 sample and 10k generations, 2) 10 samples and 1k generations, 3) 10 samples and 10k generations, and 4) 100 samples and 10k generations. The variant (1) using a HBR with only 1 sample can also be considered as a using a traditional descriptor, as the distribution cannot be captured with a single sample. However, in this case, the descriptor does not contain the initial state of the robot. The variant (2) uses the same number of evaluations as the reference algorithm, while variants (3) and (4) use respectively 10 and 100 times more evaluations.

The evaluation of each approach is made by comparing the ability of the produced BRs to execute 10k lines in random directions, starting from randomly sampled initial states. The accuracy of the BRs is defined as the median of the squared error between the requested displacements ($\mathbf{V}$) and the actual displacements executed by the robot. The experiment has been replicated 50 times for each variant.

The results show that using a stochastic descriptor reduces the number of controllers in the BRs by several orders of magnitude (at least 500 times in our experiment), while increasing the global accuracy of the architecture by at least a factor of 2.
All variants of HBRs considered, the BRs contain 1397 [1335; 1520] controllers versus 745k [719k; 766k] controllers for the BR using a traditional descriptor. 
BRs with a gigantic number of controllers raise several problems. First, it becomes particularly challenging to store and manipulate such a large collection; second, in QD-algorithms the selection pressure is often proportional to the number of solutions in the archive~\citep{cully2017quality}. BRs with 750k controllers are very likely to severely affect the selection pressure. 

\begin{figure}
\centering \includegraphics[width=\columnwidth]{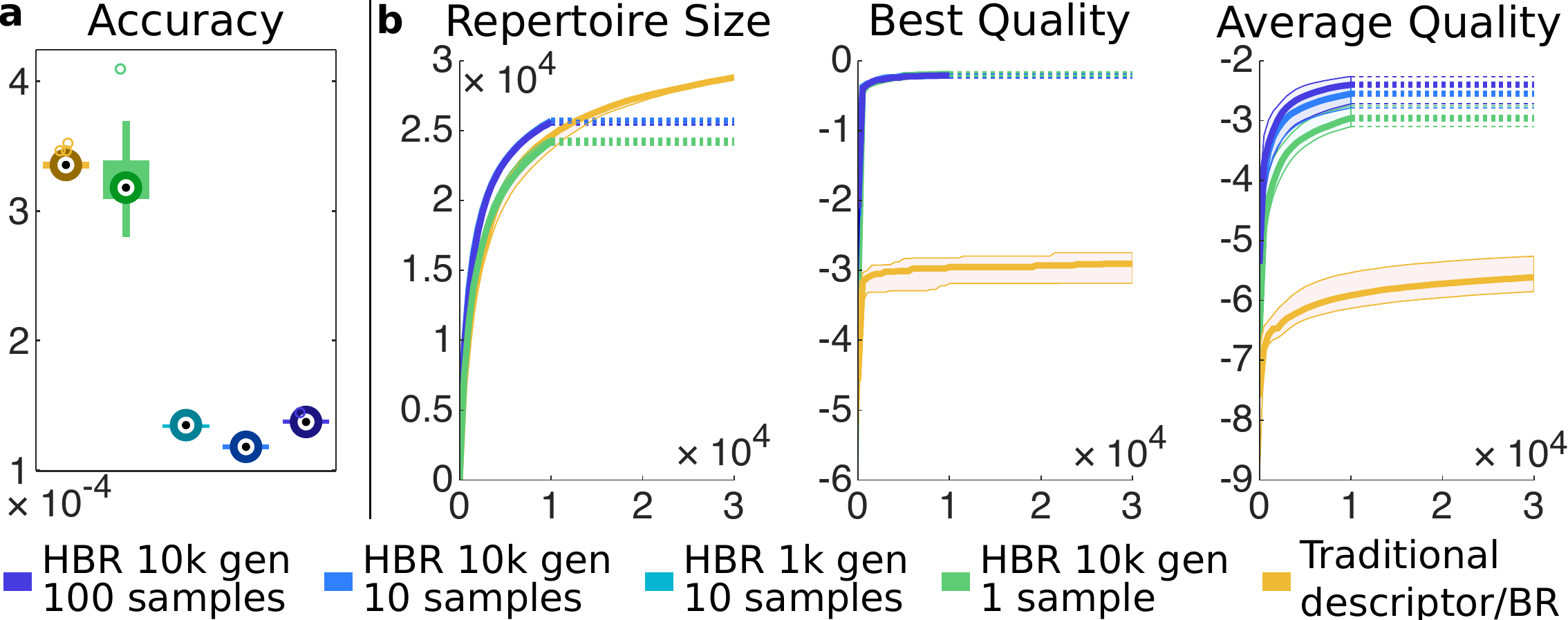}
\caption{a) Benefits of stochastic behavioral descriptors compared to traditional ones: accuracy of the layer 2 to draw 10k lines of different lengths and directions from random initial configuration (50 replications).
b) Benefits of the hierarchy in HBRs compared to traditional BRs: number of behaviors, best quality and average quality over the BR for each of the 50 replications.
}
\label{fig:res_line}
\end{figure}

In terms of accuracy, the HBRs with at least 10 samples (see Fig~\ref{fig:res_line}-a) are at least twice more accurate than the traditional BR and the HBRs with 1 sample (Wilcoxon ranksum test: $\textrm{p-values}<7.1e^{-18}$). We can also see from the results that using more than 10 samples, or make the evolution last longer, does not have an important effect on the accuracy of the BR ($<2e^{-5} \textrm{m}^2$ even though the differences are statistically significant, $p<2.6e^{-9}$).

\subsection{The benefits of the hierarchy}
In this final experiment, we evaluate the benefits of HBR compared to traditional BRs in terms of the quality of the produced behaviors. For this comparison, we compare the quality of the arcs generated by the first three layers of the architecture against those from a single traditional BR. The controllers for the reference algorithm directly control the angular position of the robot's joints. It encodes the 5 way-points trajectory of the position of the 8 joints (40 parameters). This controller is designed to have the same capabilities as the ones used with our architecture.
The fitness function of the third layer of our approach and of the compared algorithm are the same and composed of three terms as explained above (see Fig~\ref{fig:notation}-Layer3). While the two first terms promote the quality of the arcs, the last term describes the angular configuration of the robot. While this last term might appear unnecessary to draw arcs, it is crucial to guarantee that the transition between two close positions of the gripper is approximately a line. Without this constraint, the redundancy of the robot could make the transition follow an uncontrolled trajectory.

We can first observe that the hierarchy reduces the total length of the genotypes ( $8 + 2 + 3 = 13$ parameters compared to $40$ parameters). It is known that reducing the dimensionality of the search space usually reduces the difficulty of the optimization task~\citep{mataric1996challenges}.
While we can expect the HBRs to outperform traditional BRs, we can see from the results (see Fig~\ref{fig:res_line}) that the reference algorithm not only generates solutions of lower quality but that it appears to be blocked in a local optimum. Although the figure does not consider the time required to generate the two first layers of HBRs (2$*$10k generations), we can notice that the performance of the traditional BRs remains 2.3 times lower than the last layer of HBRs, even after three times more generations. We hypothesize that this phenomenon comes from the modularity of the architecture and the transfer of certain properties (like the fitness values) through the layers of the hierarchy. For instance, in this case, the first layer of the hierarchy already attempts to minimize part of the fitness function of the third layer. Thanks to that, the third layer can focus on the arc quality, without considering the motor commands sent to the robot. Similarly, it is very likely that the traditional BR is blocked in a local optimum after trying to cover the behavioral space (we can see that the covering rate is as quick as the HBRs). It might be particularly challenging to make improvements in one of the terms of the fitness function, without being detrimental to the other ones. 

We can also note that the variant with 1 sample, which can be associated to a BR with a traditional descriptor, still manages to find high performing solutions even if the average quality of the behaviors is decreased. This illustrates that the hierarchical architecture is key to improving the quality of the produced solutions. 
Among the compared variants of HBRs, we can see that evaluating more samples with the stochastic BR (10 and 100 samples) improves the average quality of produced BRs. We can hypothesize that this is due to the higher accuracy of the stochastic representation obtained by evaluating more samples.

\section{Discussion and Conclusion}
In this paper, we introduced the HBRs to extend the original concept of BRs in order to enable the generation of complex behaviors. In particular, 
the results show that the proposed architecture improves significantly the quality of the produced BRs, sometimes by several orders of magnitude, when compared to the traditional BRs. We also have shown that the proposed architecture can be coupled with an unsupervised deep neural network to automatically define the behavioral descriptor of a BR. Finally, we have illustrated the ability of the architecture to propagate fitness values through the layers of the architecture and to transfer knowledge across different robots. 

These promising results also open new research directions. For instance, we might wonder if the hierarchy can help to cross the reality gap. If the first layer of the hierarchy is made to cross the reality gap (for instance, with the Intelligent Trial and Error algorithm~\citep{cully2015robots}), does that enable the rest of the hierarchy to cross it too? Such a property might facilitate the evolution of complex behaviors for physical robots, as it very likely is easier to overcome the reality gap with simple low-level behaviors, rather than with complex high-level ones.

\appendix
\section{Parameter values}\label{ap:nn}
The QD algorithm uses the following parameter values:\\
\noindent \small{\begin{tabular}{r |ccccc}
		&Layer1&	Layer2&	Layer3&	Layer4&\\
      \cline{1-5}
archive $l$	&0.01&	0.02&	0.05&	0.1&\\
archive $\epsilon$ &	0.1	&0.1&	0.025&	0.1&\\

\multicolumn{6}{l}{$\textrm{pop\_size} = 200$, $\textrm{nb\_generations}=10000$,  no crossover}\\
\multicolumn{6}{l}{mutation: polynomial with  $\eta_m = 10$  $\eta_c = 10$, mutation rate $10\%$ }\\ 
\end{tabular}}

The architectures of the encoder and decoder are designed to be symmetrical. 
In the following, we use the notations FC, CV and DC to denote fully-connected, convolution and deconvolution layers. For example, FC(128,64) denotes two fully-connected layers with respectively 128 and 64 neurons. CV(64, 4c2s) denotes a convolution layer with 64 output feature maps, using 4x4 kernels with stride 2. DC are defined similarly as CV.
\textbf{Encoder}: (input) - CV(8, 9c1s) - maxpooling - CV(8, 3c1s) - maxpooling - FC(100,100,2) - (latent space)
\textbf{Decoder}: (latent space) - FC(100, 100, 392) - DC(8,3c1s) - unpooling - DC(8,9c1s) - unpooling - CV(1,3c1s) - (output)

To improve the contrast between the classes, the training dataset (MNIST) is reduced after the 120th epoch (for a total of 200 epochs) to the samples with a reconstruction error smaller than the current average reconstruction error. This approach enables the network to focus on samples that are simple to encode while ignoring those that might be too complicated~\citep{wang2017magan}.

\begin{acks}
This work was supported by the EU Horizon2020 Project PAL (643783-RIA).
\end{acks}

\begin{spacing}{0.9}\small
\bibliographystyle{ACM-Reference-Format}
\bibliography{biblio} 


\begin{thebibliography}{00}


\ifx \showCODEN    \undefined \def \showCODEN     #1{\unskip}     \fi
\ifx \showDOI      \undefined \def \showDOI       #1{{\tt DOI:}\penalty0{#1}\ }
  \fi
\ifx \showISBNx    \undefined \def \showISBNx     #1{\unskip}     \fi
\ifx \showISBNxiii \undefined \def \showISBNxiii  #1{\unskip}     \fi
\ifx \showISSN     \undefined \def \showISSN      #1{\unskip}     \fi
\ifx \showLCCN     \undefined \def \showLCCN      #1{\unskip}     \fi
\ifx \shownote     \undefined \def \shownote      #1{#1}          \fi
\ifx \showarticletitle \undefined \def \showarticletitle #1{#1}   \fi
\ifx \showURL      \undefined \def \showURL       #1{#1}          \fi
\providecommand\bibfield[2]{#2}
\providecommand\bibinfo[2]{#2}
\providecommand\natexlab[1]{#1}
\providecommand\showeprint[2][]{arXiv:#2}

\bibitem[\protect\citeauthoryear{Bongard, Zykov, and Lipson}{Bongard
  et~al\mbox{.}}{2006}]%
        {bongard2006resilient}
\bibfield{author}{\bibinfo{person}{J. Bongard}, \bibinfo{person}{V. Zykov},
  {and} \bibinfo{person}{H. Lipson}.} \bibinfo{year}{2006}\natexlab{}.
\newblock \showarticletitle{Resilient machines through continuous
  self-modeling}.
\newblock \bibinfo{journal}{{\em Science\/}} \bibinfo{volume}{314},
  \bibinfo{number}{5802} (\bibinfo{year}{2006}).
\newblock


\bibitem[\protect\citeauthoryear{Chatzilygeroudis, Vassiliades, and
  Mouret}{Chatzilygeroudis et~al\mbox{.}}{2018}]%
        {chatzilygeroudis2018reset}
\bibfield{author}{\bibinfo{person}{Konstantinos Chatzilygeroudis},
  \bibinfo{person}{Vassilis Vassiliades}, {and} \bibinfo{person}{Jean-Baptiste
  Mouret}.} \bibinfo{year}{2018}\natexlab{}.
\newblock \showarticletitle{Reset-free Trial-and-Error Learning for Robot
  Damage Recovery}.
\newblock \bibinfo{journal}{{\em Robotics and Autonomous Systems\/}}
  \bibinfo{volume}{100} (\bibinfo{year}{2018}), \bibinfo{pages}{236--250}.
\newblock


\bibitem[\protect\citeauthoryear{Cully, Clune, Tarapore, and Mouret}{Cully
  et~al\mbox{.}}{2015}]%
        {cully2015robots}
\bibfield{author}{\bibinfo{person}{Antoine Cully}, \bibinfo{person}{Jeff
  Clune}, \bibinfo{person}{Danesh Tarapore}, {and}
  \bibinfo{person}{Jean-Baptiste Mouret}.} \bibinfo{year}{2015}\natexlab{}.
\newblock \showarticletitle{Robots that can adapt like animals}.
\newblock \bibinfo{journal}{{\em Nature\/}} \bibinfo{volume}{521},
  \bibinfo{number}{7553} (\bibinfo{year}{2015}), \bibinfo{pages}{503--507}.
\newblock


\bibitem[\protect\citeauthoryear{Cully and Demiris}{Cully and Demiris}{2017}]%
        {cully2017quality}
\bibfield{author}{\bibinfo{person}{Antoine Cully} {and}
  \bibinfo{person}{Yiannis Demiris}.} \bibinfo{year}{2017}\natexlab{}.
\newblock \showarticletitle{Quality and Diversity Optimization: A Unifying
  Modular Framework}.
\newblock \bibinfo{journal}{{\em IEEE Trans. on Evolutionary Computation\/}}
  (\bibinfo{year}{2017}).
\newblock


\bibitem[\protect\citeauthoryear{Cully and Mouret}{Cully and Mouret}{2013}]%
        {cully2013behavioral}
\bibfield{author}{\bibinfo{person}{Antoine Cully} {and}
  \bibinfo{person}{Jean-Baptiste Mouret}.} \bibinfo{year}{2013}\natexlab{}.
\newblock \showarticletitle{Behavioral repertoire learning in robotics}. In
  \bibinfo{booktitle}{{\em Proceedings of the 15th annual conference on Genetic
  and Evolutionary Computation}}. ACM, \bibinfo{pages}{175--182}.
\newblock


\bibitem[\protect\citeauthoryear{Cully and Mouret}{Cully and Mouret}{2015}]%
        {cully2015evolving}
\bibfield{author}{\bibinfo{person}{Antoine Cully} {and}
  \bibinfo{person}{Jean-Baptiste Mouret}.} \bibinfo{year}{2015}\natexlab{}.
\newblock \showarticletitle{Evolving a behavioral repertoire for a walking
  robot}.
\newblock \bibinfo{journal}{{\em Evolutionary Computation\/}}
  (\bibinfo{year}{2015}).
\newblock


\bibitem[\protect\citeauthoryear{Doncieux, Bredeche, Mouret, and
  Eiben}{Doncieux et~al\mbox{.}}{2015}]%
        {doncieux2015evolutionary}
\bibfield{author}{\bibinfo{person}{Stephane Doncieux}, \bibinfo{person}{Nicolas
  Bredeche}, \bibinfo{person}{Jean-Baptiste Mouret}, {and}
  \bibinfo{person}{Agoston E~Gusz Eiben}.} \bibinfo{year}{2015}\natexlab{}.
\newblock \showarticletitle{Evolutionary robotics: what, why, and where to}.
\newblock \bibinfo{journal}{{\em Frontiers in Robotics and AI\/}}
  \bibinfo{volume}{2} (\bibinfo{year}{2015}), \bibinfo{pages}{4}.
\newblock


\bibitem[\protect\citeauthoryear{Duarte, Gomes, Oliveira, and
  Christensen}{Duarte et~al\mbox{.}}{2017}]%
        {duarte2017evolution}
\bibfield{author}{\bibinfo{person}{Miguel Duarte}, \bibinfo{person}{Jorge
  Gomes}, \bibinfo{person}{Sancho~Moura Oliveira}, {and}
  \bibinfo{person}{Anders~Lyhne Christensen}.} \bibinfo{year}{2017}\natexlab{}.
\newblock \showarticletitle{Evolution of repertoire-based control for robots
  with complex locomotor systems}.
\newblock \bibinfo{journal}{{\em IEEE Transactions on Evolutionary
  Computation\/}} (\bibinfo{year}{2017}).
\newblock


\bibitem[\protect\citeauthoryear{Eiben and Smith}{Eiben and Smith}{2015}]%
        {eiben2015evolutionary}
\bibfield{author}{\bibinfo{person}{Agoston~E Eiben} {and} \bibinfo{person}{Jim
  Smith}.} \bibinfo{year}{2015}\natexlab{}.
\newblock \showarticletitle{From evolutionary computation to the evolution of
  things}.
\newblock \bibinfo{journal}{{\em Nature\/}} \bibinfo{volume}{521},
  \bibinfo{number}{7553} (\bibinfo{year}{2015}), \bibinfo{pages}{476--482}.
\newblock


\bibitem[\protect\citeauthoryear{Gaier, Asteroth, and Mouret}{Gaier
  et~al\mbox{.}}{2017}]%
        {gaier2017data}
\bibfield{author}{\bibinfo{person}{Adam Gaier}, \bibinfo{person}{Alexander
  Asteroth}, {and} \bibinfo{person}{Jean-Baptiste Mouret}.}
  \bibinfo{year}{2017}\natexlab{}.
\newblock \showarticletitle{Data-Efficient Exploration, Optimization, and
  Modeling of Diverse Designs through Surrogate-Assisted Illumination}. In
  \bibinfo{booktitle}{{\em Genetic and Evolutionary Computation Conference}}.
\newblock


\bibitem[\protect\citeauthoryear{Gomez and Miikkulainen}{Gomez and
  Miikkulainen}{1997}]%
        {gomez1997incremental}
\bibfield{author}{\bibinfo{person}{Faustino Gomez} {and} \bibinfo{person}{Risto
  Miikkulainen}.} \bibinfo{year}{1997}\natexlab{}.
\newblock \showarticletitle{Incremental evolution of complex general behavior}.
\newblock \bibinfo{journal}{{\em Adaptive Behavior\/}} \bibinfo{volume}{5},
  \bibinfo{number}{3-4} (\bibinfo{year}{1997}), \bibinfo{pages}{317--342}.
\newblock


\bibitem[\protect\citeauthoryear{Goodfellow, Bengio, and Courville}{Goodfellow
  et~al\mbox{.}}{2016}]%
        {Goodfellow-et-al-2016}
\bibfield{author}{\bibinfo{person}{Ian Goodfellow}, \bibinfo{person}{Yoshua
  Bengio}, {and} \bibinfo{person}{Aaron Courville}.}
  \bibinfo{year}{2016}\natexlab{}.
\newblock \bibinfo{booktitle}{{\em Deep Learning}}.
\newblock \bibinfo{publisher}{MIT Press}.
\newblock
\newblock
\shownote{\url{http://www.deeplearningbook.org}.}


\bibitem[\protect\citeauthoryear{Gravina, Liapis, and Yannakakis}{Gravina
  et~al\mbox{.}}{2016}]%
        {gravina2016surprise}
\bibfield{author}{\bibinfo{person}{Daniele Gravina}, \bibinfo{person}{Antonios
  Liapis}, {and} \bibinfo{person}{Georgios Yannakakis}.}
  \bibinfo{year}{2016}\natexlab{}.
\newblock \showarticletitle{Surprise search: Beyond objectives and novelty}. In
  \bibinfo{booktitle}{{\em Proceedings of the 2016 on Genetic and Evolutionary
  Computation Conference}}. ACM, \bibinfo{pages}{677--684}.
\newblock


\bibitem[\protect\citeauthoryear{Larsen and Hansen}{Larsen and Hansen}{2005}]%
        {larsen2005evolving}
\bibfield{author}{\bibinfo{person}{Tobias Larsen} {and}
  \bibinfo{person}{S{\o}ren~Tranberg Hansen}.} \bibinfo{year}{2005}\natexlab{}.
\newblock \showarticletitle{Evolving composite robot behaviour-a modular
  architecture}. In \bibinfo{booktitle}{{\em Robot Motion and Control, 2005.
  RoMoCo'05. Proceedings of the Fifth International Workshop on}}. IEEE,
  \bibinfo{pages}{271--276}.
\newblock


\bibitem[\protect\citeauthoryear{LeCun, Bengio, and Hinton}{LeCun
  et~al\mbox{.}}{2015}]%
        {lecun2015deep}
\bibfield{author}{\bibinfo{person}{Yann LeCun}, \bibinfo{person}{Yoshua
  Bengio}, {and} \bibinfo{person}{Geoffrey Hinton}.}
  \bibinfo{year}{2015}\natexlab{}.
\newblock \showarticletitle{Deep learning}.
\newblock \bibinfo{journal}{{\em Nature\/}} \bibinfo{volume}{521},
  \bibinfo{number}{7553} (\bibinfo{year}{2015}), \bibinfo{pages}{436--444}.
\newblock


\bibitem[\protect\citeauthoryear{Lehman and Stanley}{Lehman and
  Stanley}{2011}]%
        {lehman2011evolving}
\bibfield{author}{\bibinfo{person}{Joel Lehman} {and}
  \bibinfo{person}{Kenneth~O Stanley}.} \bibinfo{year}{2011}\natexlab{}.
\newblock \showarticletitle{Evolving a diversity of virtual creatures through
  novelty search and local competition}. In \bibinfo{booktitle}{{\em
  Proceedings of the 13th annual conference on Genetic and Evolutionary
  Computation}}. ACM, \bibinfo{pages}{211--218}.
\newblock


\bibitem[\protect\citeauthoryear{Lin and Vitter}{Lin and Vitter}{1992}]%
        {lin1992approximation}
\bibfield{author}{\bibinfo{person}{Jyh-Han Lin} {and}
  \bibinfo{person}{Jeffrey~Scott Vitter}.} \bibinfo{year}{1992}\natexlab{}.
\newblock \showarticletitle{Approximation algorithms for geometric median
  problems}.
\newblock \bibinfo{journal}{{\it Inform. Process. Lett.}} \bibinfo{volume}{44},
  \bibinfo{number}{5} (\bibinfo{year}{1992}), \bibinfo{pages}{245--249}.
\newblock


\bibitem[\protect\citeauthoryear{Lipson and Pollack}{Lipson and
  Pollack}{2000}]%
        {lipson2000automatic}
\bibfield{author}{\bibinfo{person}{Hod Lipson} {and} \bibinfo{person}{Jordan~B
  Pollack}.} \bibinfo{year}{2000}\natexlab{}.
\newblock \showarticletitle{Automatic design and manufacture of robotic
  lifeforms}.
\newblock \bibinfo{journal}{{\em Nature\/}} \bibinfo{volume}{406},
  \bibinfo{number}{6799} (\bibinfo{year}{2000}), \bibinfo{pages}{974--978}.
\newblock


\bibitem[\protect\citeauthoryear{Matari{\'c} and Cliff}{Matari{\'c} and
  Cliff}{1996}]%
        {mataric1996challenges}
\bibfield{author}{\bibinfo{person}{Maja Matari{\'c}} {and}
  \bibinfo{person}{Dave Cliff}.} \bibinfo{year}{1996}\natexlab{}.
\newblock \showarticletitle{Challenges in evolving controllers for physical
  robots}.
\newblock \bibinfo{journal}{{\em Robotics and autonomous systems\/}}
  \bibinfo{volume}{19}, \bibinfo{number}{1} (\bibinfo{year}{1996}),
  \bibinfo{pages}{67--83}.
\newblock


\bibitem[\protect\citeauthoryear{Mengistu, Lehman, and Clune}{Mengistu
  et~al\mbox{.}}{2016}]%
        {mengistu2016evolvability}
\bibfield{author}{\bibinfo{person}{Henok Mengistu}, \bibinfo{person}{Joel
  Lehman}, {and} \bibinfo{person}{Jeff Clune}.}
  \bibinfo{year}{2016}\natexlab{}.
\newblock \showarticletitle{Evolvability search: directly selecting for
  evolvability in order to study and produce it}. In \bibinfo{booktitle}{{\em
  Proceedings of the 2016 on Genetic and Evolutionary Computation Conference}}.
  ACM, \bibinfo{pages}{141--148}.
\newblock


\bibitem[\protect\citeauthoryear{Mouret and Clune}{Mouret and Clune}{2015}]%
        {mouret2015illuminating}
\bibfield{author}{\bibinfo{person}{Jean-Baptiste Mouret} {and}
  \bibinfo{person}{Jeff Clune}.} \bibinfo{year}{2015}\natexlab{}.
\newblock \showarticletitle{Illuminating search spaces by mapping elites}.
\newblock \bibinfo{journal}{{\em arXiv preprint arXiv:1504.04909\/}}
  (\bibinfo{year}{2015}).
\newblock


\bibitem[\protect\citeauthoryear{Mouret and Doncieux}{Mouret and
  Doncieux}{2008}]%
        {mouret2008incremental}
\bibfield{author}{\bibinfo{person}{Jean-Baptiste Mouret} {and}
  \bibinfo{person}{St{\'e}phane Doncieux}.} \bibinfo{year}{2008}\natexlab{}.
\newblock \showarticletitle{Incremental evolution of animats' behaviors as a
  multi-objective optimization}. In \bibinfo{booktitle}{{\em International
  Conference on Simulation of Adaptive Behavior}}. Springer,
  \bibinfo{pages}{210--219}.
\newblock


\bibitem[\protect\citeauthoryear{Nguyen, Yosinski, and Clune}{Nguyen
  et~al\mbox{.}}{2015a}]%
        {nguyen2015deep}
\bibfield{author}{\bibinfo{person}{Anh Nguyen}, \bibinfo{person}{Jason
  Yosinski}, {and} \bibinfo{person}{Jeff Clune}.}
  \bibinfo{year}{2015}\natexlab{a}.
\newblock \showarticletitle{Deep neural networks are easily fooled: High
  confidence predictions for unrecognizable images}. In
  \bibinfo{booktitle}{{\em Conference on Computer Vision and Pattern
  Recognition}}. IEEE.
\newblock


\bibitem[\protect\citeauthoryear{Nguyen, Yosinski, and Clune}{Nguyen
  et~al\mbox{.}}{2016}]%
        {nguyen2016understanding}
\bibfield{author}{\bibinfo{person}{Anh Nguyen}, \bibinfo{person}{Jason
  Yosinski}, {and} \bibinfo{person}{Jeff Clune}.}
  \bibinfo{year}{2016}\natexlab{}.
\newblock \showarticletitle{Understanding innovation engines: Automated
  creativity and improved stochastic optimization via deep learning}.
\newblock \bibinfo{journal}{{\em Evolutionary computation\/}}
  \bibinfo{volume}{24}, \bibinfo{number}{3} (\bibinfo{year}{2016}),
  \bibinfo{pages}{545--572}.
\newblock


\bibitem[\protect\citeauthoryear{Nguyen, Yosinski, and Clune}{Nguyen
  et~al\mbox{.}}{2015b}]%
        {nguyen2015innovation}
\bibfield{author}{\bibinfo{person}{Anh~Mai Nguyen}, \bibinfo{person}{Jason
  Yosinski}, {and} \bibinfo{person}{Jeff Clune}.}
  \bibinfo{year}{2015}\natexlab{b}.
\newblock \showarticletitle{Innovation engines: Automated creativity and
  improved stochastic optimization via deep learning}. In
  \bibinfo{booktitle}{{\em Proceedings of the 2015 Annual Conference on Genetic
  and Evolutionary Computation}}. ACM, \bibinfo{pages}{959--966}.
\newblock


\bibitem[\protect\citeauthoryear{Nolfi and Parisi}{Nolfi and Parisi}{1995}]%
        {nolfi1995evolving}
\bibfield{author}{\bibinfo{person}{Stefano Nolfi} {and}
  \bibinfo{person}{Domenico Parisi}.} \bibinfo{year}{1995}\natexlab{}.
\newblock \showarticletitle{Evolving non-trivial behaviors on real robots: an
  autonomous robot that picks up objects}. In \bibinfo{booktitle}{{\em Congress
  of the Italian Association for Artificial Intelligence}}. Springer,
  \bibinfo{pages}{243--254}.
\newblock


\bibitem[\protect\citeauthoryear{Pugh, Soros, Szerlip, and Stanley}{Pugh
  et~al\mbox{.}}{2015}]%
        {pugh2015confronting}
\bibfield{author}{\bibinfo{person}{Justin~K Pugh}, \bibinfo{person}{LB Soros},
  \bibinfo{person}{Paul~A Szerlip}, {and} \bibinfo{person}{Kenneth~O Stanley}.}
  \bibinfo{year}{2015}\natexlab{}.
\newblock \showarticletitle{Confronting the challenge of quality diversity}. In
  \bibinfo{booktitle}{{\em Proceedings of the 2015 on Genetic and Evolutionary
  Computation Conference}}. ACM, \bibinfo{pages}{967--974}.
\newblock


\bibitem[\protect\citeauthoryear{Pugh, Soros, and Stanley}{Pugh
  et~al\mbox{.}}{2016}]%
        {pugh2016quality}
\bibfield{author}{\bibinfo{person}{Justin~K Pugh}, \bibinfo{person}{Lisa~B
  Soros}, {and} \bibinfo{person}{Kenneth~O Stanley}.}
  \bibinfo{year}{2016}\natexlab{}.
\newblock \showarticletitle{Quality Diversity: A New Frontier for Evolutionary
  Computation}.
\newblock \bibinfo{journal}{{\em Frontiers in Robotics and AI\/}}
  (\bibinfo{year}{2016}).
\newblock


\bibitem[\protect\citeauthoryear{Smith, Tokarchuk, and Wiggins}{Smith
  et~al\mbox{.}}{2016}]%
        {smith2016rapid}
\bibfield{author}{\bibinfo{person}{Davy Smith}, \bibinfo{person}{Laurissa
  Tokarchuk}, {and} \bibinfo{person}{Geraint Wiggins}.}
  \bibinfo{year}{2016}\natexlab{}.
\newblock \showarticletitle{Rapid Phenotypic Landscape Exploration Through
  Hierarchical Spatial Partitioning}. In \bibinfo{booktitle}{{\em International
  Conference on Parallel Problem Solving from Nature}}. Springer,
  \bibinfo{pages}{911--920}.
\newblock


\bibitem[\protect\citeauthoryear{Stanley, D'Ambrosio, and Gauci}{Stanley
  et~al\mbox{.}}{2009}]%
        {stanley2009hypercube}
\bibfield{author}{\bibinfo{person}{Kenneth~O Stanley}, \bibinfo{person}{David~B
  D'Ambrosio}, {and} \bibinfo{person}{Jason Gauci}.}
  \bibinfo{year}{2009}\natexlab{}.
\newblock \showarticletitle{A hypercube-based encoding for evolving large-scale
  neural networks}.
\newblock \bibinfo{journal}{{\em Artificial life\/}} \bibinfo{volume}{15},
  \bibinfo{number}{2} (\bibinfo{year}{2009}), \bibinfo{pages}{185--212}.
\newblock


\bibitem[\protect\citeauthoryear{Vassiliades, Chatzilygeroudis, and
  Mouret}{Vassiliades et~al\mbox{.}}{2017}]%
        {vassiliades2017using}
\bibfield{author}{\bibinfo{person}{Vassilis Vassiliades},
  \bibinfo{person}{Konstantinos Chatzilygeroudis}, {and}
  \bibinfo{person}{Jean-Baptiste Mouret}.} \bibinfo{year}{2017}\natexlab{}.
\newblock \showarticletitle{Using centroidal voronoi tessellations to scale up
  the multi-dimensional archive of phenotypic elites algorithm}.
\newblock \bibinfo{journal}{{\em IEEE Transactions on Evolutionary
  Computation\/}} (\bibinfo{year}{2017}).
\newblock


\bibitem[\protect\citeauthoryear{Wagner, Pavlicev, and Cheverud}{Wagner
  et~al\mbox{.}}{2007}]%
        {wagner2007road}
\bibfield{author}{\bibinfo{person}{G{\"u}nter~P Wagner},
  \bibinfo{person}{Mihaela Pavlicev}, {and} \bibinfo{person}{James~M
  Cheverud}.} \bibinfo{year}{2007}\natexlab{}.
\newblock \showarticletitle{The road to modularity}.
\newblock \bibinfo{journal}{{\em Nature Reviews Genetics\/}}
  \bibinfo{volume}{8}, \bibinfo{number}{12} (\bibinfo{year}{2007}),
  \bibinfo{pages}{921--931}.
\newblock


\bibitem[\protect\citeauthoryear{Wang, Cully, Chang, and Demiris}{Wang
  et~al\mbox{.}}{2017}]%
        {wang2017magan}
\bibfield{author}{\bibinfo{person}{Ruohan Wang}, \bibinfo{person}{Antoine
  Cully}, \bibinfo{person}{Hyung~Jin Chang}, {and} \bibinfo{person}{Yiannis
  Demiris}.} \bibinfo{year}{2017}\natexlab{}.
\newblock \showarticletitle{MAGAN: Margin Adaptation for Generative Adversarial
  Networks}.
\newblock \bibinfo{journal}{{\em arXiv preprint arXiv:1704.03817\/}}
  (\bibinfo{year}{2017}).
\newblock


\bibitem[\protect\citeauthoryear{Zhou, Zemanov{\'a}, Zamora, Hilgetag, and
  Kurths}{Zhou et~al\mbox{.}}{2006}]%
        {zhou2006hierarchical}
\bibfield{author}{\bibinfo{person}{Changsong Zhou}, \bibinfo{person}{Lucia
  Zemanov{\'a}}, \bibinfo{person}{Gorka Zamora}, \bibinfo{person}{Claus~C
  Hilgetag}, {and} \bibinfo{person}{J{\"u}rgen Kurths}.}
  \bibinfo{year}{2006}\natexlab{}.
\newblock \showarticletitle{Hierarchical organization unveiled by functional
  connectivity in complex brain networks}.
\newblock \bibinfo{journal}{{\em Physical review letters\/}}
  \bibinfo{volume}{97}, \bibinfo{number}{23} (\bibinfo{year}{2006}),
  \bibinfo{pages}{238103}.
\newblock


\end{thebibliography}
\end{spacing}

\end{document}